\documentclass[journal]{IEEEtai}

\usepackage[colorlinks,urlcolor=blue,linkcolor=blue,citecolor=blue]{hyperref}
\usepackage{cite}
\usepackage{color,array}
\usepackage{booktabs}
\usepackage{graphicx}
\usepackage{svg}
\usepackage{subfigure}
\usepackage{pgfplots}
\usepackage{multirow}
\usepackage{amsmath}
\usepackage{xcolor,soul}
\usepackage[most]{tcolorbox}
\usepackage{pdfcomment}
\usepackage[T1]{fontenc}

\usepackage{caption}
\newcommand{\citet}[1]{\cite{#1}}

\begin{document}

%\title{LibriSQA: Advancing Free-form and Open-ended Spoken Question Answering with a Novel Dataset and Framework } 
\title{LibriSQA: A Novel Dataset and Framework for Spoken Question Answering with Large Language Models} 

\author{Zihan Zhao, \IEEEmembership{Member, IEEE}, Yiyang Jiang, Heyang Liu, Yu Wang, \IEEEmembership{Member, IEEE}, Yanfeng Wang, \IEEEmembership{Member, IEEE}
%\thanks{This paragraph of the first footnote will contain the date on which you submitted your paper for review. It will also contain support information, including sponsor and financial support acknowledgment. For example, ``This work was supported in part by the U.S. Department of Commerce under Grant BS123456.'' }
\thanks{This work was supported by National Key R\&D Program of China (No.2022ZD0162101), Shanghai Science and Technology Committee (No.21511101100) and Shanghai Key Laboratory of Digital Media Processing and Transmission (STCSM 22DZ2229005). (\emph{Corresponding author: Yu Wang, Yanfeng Wang.})}
\thanks{Zihan Zhao, Yanfeng Wang, Yu Wang are with the Cooperative Medianet Innovation Center, Shanghai Jiao Tong University, Shanghai, China and Shanghai AI Laboratory, Shanghai, China (e-mail: {zihanzhao, wangyanfeng, yuwangsjtu}@sjtu.edu.cn).}
\thanks{Yiyang Jiang is with Shanghai AI Laboratory, Shanghai, China (e-mail: jiangyiyang@pjlab.org.cn).}
\thanks{Heyang Liu is with the Cooperative Medianet Innovation Center, Shanghai Jiao Tong University, Shanghai, China (e-mail: liuheyang@sjtu.edu.cn).}
\thanks{This work was completed when Zihan Zhao and Yiyang Jiang were interns at Shanghai AI Laboratory.}
}

%\markboth{Journal of IEEE Transactions on Artificial Intelligence, Vol. 00, No. 0, Month 2020}
%{First A. Author \MakeLowercase{\textit{et al.}}: Bare Demo of IEEEtai.cls for IEEE Journals of IEEE Transactions on Artificial Intelligence}

\maketitle

\begin{abstract}
While Large Language Models (LLMs) have demonstrated impressive performance across various domains and tasks, they still struggle with multimodal tasks, particularly the Spoken Question Answering (SQA) task, which requires precise alignment and deep interaction between speech and text. In this paper, we address the SQA challenge by curating a novel free-form and open-ended SQA dataset, LibriSQA, which is composed of 214k SQA pairs covering a wide range of topics. It consists of two parts. Part I is designed for natural conversational formats and Part II focuses on multiple-choice questions with answers and analytical segments.
Considering the limited availability of speech-text LLMs, we propose a lightweight, end-to-end framework to perform the SQA task on the LibriSQA dataset, achieving significant results. By transforming Automatic Speech Recognition (ASR) into the SQA format, we further demonstrate the framework's capability in handling ASR tasks. Our empirical findings support the idea that LLMs can effectively align and comprehend speech information, paving the way for the development of universal multimodal LLMs. Our LibriSQA dataset can be found at https://github.com/ZihanZhaoSJTU/LibriSQA.
\end{abstract}

\begin{IEEEImpStatement}

%The impact statement should not exceed 150 words. This section offers an example that is expanded to have only and just 150 words to demonstrate the point. Here is an example on how to write an appropriate impact statement: Chatbots are a popular technology in online interaction. They reduce the load on human support teams and offer continuous 24-7 support to customers. However, recent usability research has demonstrated that 30\% of customers are unhappy with current chatbots due to their poor conversational capabilities and inability to emotionally engage customers. The natural language algorithms we introduce in this paper overcame these limitations. With a significant increase in user satisfaction to 92\% after adopting our algorithms, the technology is ready to support users in a wide variety of applications including government front shops, automatic tellers, and the gaming industry. It could offer an alternative way of interaction for some physically disable users.
This research addresses a critical gap in the capabilities of Large Language Models (LLMs) concerning multimodal tasks, particularly focusing on the Spoken Question Answering (SQA) task which demands intricate alignment and deep interaction between speech and text. By collecting a comprehensive and diverse LibriSQA dataset and introducing a novel, lightweight end-to-end framework, we have made noteworthy advancements in performing SQA tasks, witnessing substantial results that achieve 71.1\% accuracy in four-option questions with only about 2\% of the trainable parameters compared to other state-of-the-art approaches. Our lightweight framework also achieves promising results on the ASR task, significantly improving training and inference speed while reducing resource usage. These developments not only underline the enhanced ability of LLMs in aligning and understanding multimodal information but also signify a pivotal step towards the evolution of universal multimodal LLMs, contributing to the field’s progress.
\end{IEEEImpStatement}


\section{Introduction}
\label{intro}

\IEEEPARstart{L}{arge} language models (LLMs) have witnessed rapid advancements in recent times, exhibiting considerable superiority in text generation over preceding methodologies \cite{gpt3,chatgpt,llama,gpt4,llama2}. Despite these achievements, current LLMs demonstrate limited proficiency in processing and comprehending multimodal inputs. While efforts have been made to integrate visual inputs with LLMs, resulting in notable accomplishments in various tasks \cite{multimodalgpt,llamaadapter,pmcvqa}, the combination of speech and text modalities within LLMs remains relatively under-explored. Previous models with speech and text mostly focused on classification problems such as multimodal emotion recognition \cite{m1,m2,Auxiliarytask,GCF2-Net}. For spoken language understanding, early works used a cascade approach \cite{cascadeslu}, and later works utilized the end-to-end approach \cite{e2eslu}. Some works also used speech and text as multimodal inputs \cite{slu1,tie}. In addition, regression tasks like score prediction or uncertainty estimation were widely explored \cite{spontaneousse, uncerest}. In this paper, we focus on the interaction of speech and text modalities with LLMs to fill the gaps. Some recent works have proposed research in this area. Previous efforts by Huang et al. \cite{audiogpt} have incorporated external Automatic Speech Recognition (ASR) and Text-To-Speech (TTS) systems to bridge the gap between speech and text. However, reliance on these external mechanisms not only increases resource requirements but also introduces potential inaccuracies due to the inherent limitations of these systems. In pursuit of an end-to-end approach, Zhang et al. \cite{speechgpt} devised a multimodal instruction dataset and utilized it to train LLaMA, subsequently facilitating the model to be able to complete both ASR, TTS and spoken dialogue without the dependency on auxiliary ASR and TTS systems. Another noteworthy attempt is by Rubenstein et al. \cite{audiopalm}, who adapted a decoder-only Transformer to handle multimodal translation, TTS, and ASR tasks. 
%While these innovations signify meaningful strides towards harmonizing speech and text interactions, a palpable void emerges in their capability to engage in unconstrained, open-ended question-answering (QA) of spoken content - an indispensable facet for next-generation conversational AI. Such challenges, encapsulated within the Spoken Question Answering (SQA) paradigm, underscore the essence of this work.
Although these innovations mark significant progress in integrating speech and text interactions, there is still a noticeable gap in their ability to engage in unrestricted, open-ended Question Answering (QA) of spoken content. This capability is essential for next-generation conversational artificial intelligence. These challenges, encompassed within the Spoken Question Answering (SQA) paradigm, highlight the significance of this work.
%We posit that empowering LLMs with adeptness in SQA is paramount for the realization of truly multimodal conversational AI systems.

 %Existing SQA dataset inputs typically use speech as background information, use text as the question, and then either find a word or phrase in the background speech or answer yes or no to the question \cite{spokensquad,odsqa}. The common way to find answer words or phrases is to let the model predict a start point and an end point in the background speech. The TOEFL dataset collected by Tseng et al. was another kind of SQA dataset that was in multiple-choice form and had only an option as answer \cite{toefl}. However, we think it would be more natural to answer the question by using a free-form and open-ended answer. Spoken-CoQA is the first SQA dataset in the form of conversations, and it also has free-form answers, which is probably the closest to the dataset we want \cite{spokencoqa}. However, it has the following problems: 1) the speech is not real speech but generated by TTS, which is not close to the real situation, 2) each of these voices is several minutes long, which makes it hard to be used by LLMs if we don't consider compression, and our main concern is the alignment of speech and text on a large model, how to compress the speech without affecting the effect of SQA is not our goal now, 3) although it is a free-form Q&A, the answers are usually not complete sentences but one or two words, which can not reflect the powerful text generation capability of LLMs.
 
\begin{table*}[]
\centering % 居中表格
  \caption{Comparison between different SQA datasets}
  \label{datasetcompare}
\begin{tabular}{l|ccccc}
\toprule
\textbf{Dataset} & \textbf{No. of Speech Samples} & \textbf{Language} & \textbf{Speech Type} & \textbf{Question Type} & \textbf{Answer Type}   \\ \midrule
Spoken SQuAD \cite{spokensquad}     & 21k                     & English           & Synthetic            & Natural Question       & Time Span              \\
ODSQA \cite{odsqa}            & 9k                      & Chinese           & Authentic            & Natural Question       & Time Span              \\
TOEFL \cite{toefl}            & 0.9k                       & English           & Authentic            & Question+Multi-choice  & Option Label           \\
Spoken-CoQA \cite{spokencoqa}      & 8k                      & English           & Synthetic            & Natural Question       & Free-form              \\ \midrule
LibriSQA Part I  & 107k                    & English           & Authentic            & Natural Question       & Free-form              \\
LibriSQA Part II & 107k                    & English           & Authentic            & Question+Multi-choice  & Option Label\&Analysis \\ \bottomrule
\end{tabular}
\end{table*}

Existing SQA datasets conventionally use speech as the contextual background, alongside textual questions. The typical paradigm then requires models to identify specific word or phrase locations within the spoken context or to yield binary affirmations or negations in response to the questions \cite{spokensquad,odsqa}. This kind of problem is often realized by having the model predict starting and ending points within the background speech. An alternative form is seen in the TOEFL dataset collected by Tseng et al. \cite{toefl}, where the SQA is presented through questions with several options and the answer option. 
Nevertheless, we advocate for free-form, open-ended QA as this is more intuitive and natural. The Spoken-CoQA dataset behaved as a conversational SQA with free-form QA and emerged as the dataset that resembles our envisioned one \cite{spokencoqa}. This dataset, however, was not without its limitations: 1) The speech content was synthesized through TTS, diverging from authentic speech scenarios; 2) The speech segments span multiple minutes, presenting challenges for seamless integration with LLMs, especially without resorting to compression techniques. It is worth noting that our primary focus revolves around optimizing speech-text alignment in LLMs and compressing speech without compromising SQA efficacy is not our main focus; 3) {While the Spoken-CoQA model adopts a free-form question-answering approach, the responses typically consist of simple one or two-word answers extracted directly from the text. Therefore, it remains feasible to predict answers by identifying specific time spans in the context }\cite{spokencoqa}, {which unfortunately does not take full advantage of the advanced text generation capabilities of LLMs.}

{Due to the limitations of the existing SQA dataset, current methodologies for SQA typically prioritize predicting time spans or option labels over employing generative text }\cite{speechbert,splat,threetasksqa,spokencoqa}. {As a result, it becomes evident that the existing SQA datasets and methodologies are not ideally suited for both research and deployment with LLMs. To address these issues, we have made the following contributions:}

\begin{itemize}
%\item We introduce the LibriSQA dataset. To the best of our knowledge, it represents the first SQA dataset encompassing free-form and open-ended question-answer pairs specifically optimized for LLMs, a paradigm shift from the traditional approach of predicting temporal spans.
\item In this paper we introduce the  LibriSQA dataset. To the best of our knowledge, it is the first SQA dataset that includes free-form and open-ended question-answer pairs specifically optimized for LLMs. This represents a paradigm shift from the traditional approach of predicting temporal spans.

%\item We propose a lightweight, end-to-end framework that integrates both speech and text into LLMs, obviating the need for ASR modules. This underscores the intrinsic capability of LLMs to interpret and process speech autonomously, without reliance on external utilities.
\item We propose a lightweight, end-to-end framework that seamlessly integrates both speech and text into LLMs, eliminating the need for ASR modules. This highlights the inherent capability of LLMs to independently interpret and process speech, without relying on external utilities.

%\item Employing our novel framework, we fine-tune LLMs utilizing both reformed LibriSpeech and our LibriSQA. The results attest to the efficacy of our framework across ASR and SQA tasks, facilitating deeper interactions between speech and text within LLMs.
\item By utilizing our innovative framework, we successfully fine-tune LLMs using both the reformed LibriSpeech dataset and our LibriSQA dataset. The results unequivocally demonstrate the effectiveness of our framework across ASR and SQA tasks, enabling deeper interactions between speech and text within LLMs.

\end{itemize}

\begin{figure}
\centerline{\includegraphics[width=22pc]{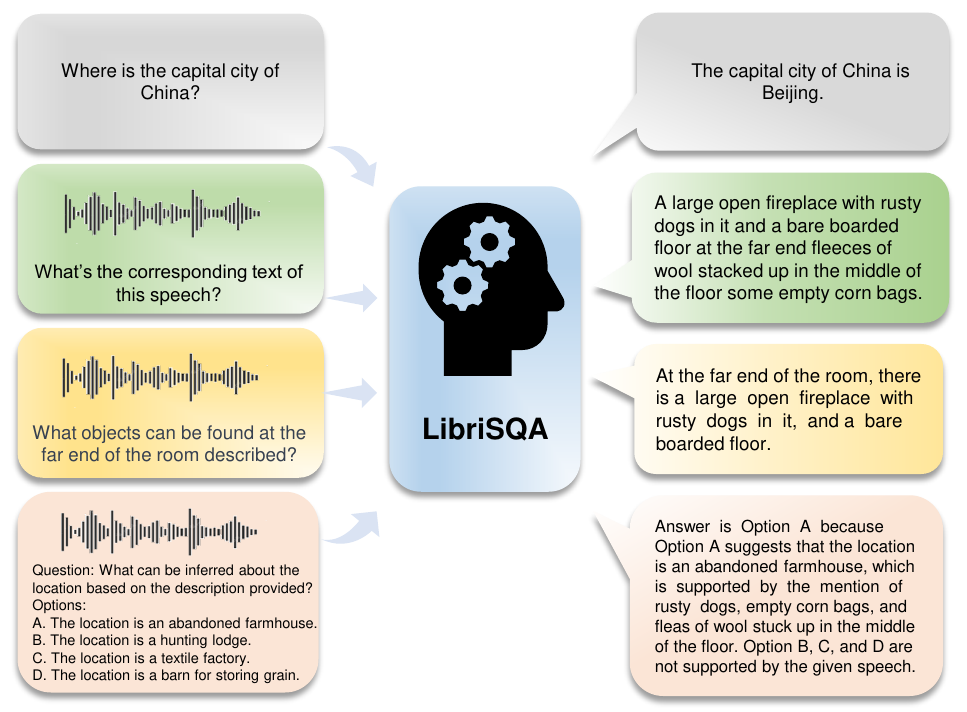}}

\caption{Four application scenarios of our framework are described from the top to bottom. The first scenario involves a text-only Question Answering (QA) modality. The remaining scenarios are multimodal QAs, where the background information is provided through speech and the questions are presented in textual modality. The third and fourth scenarios are taken from LibriSQA Part I and Part II respectively. In the example shown here, the same speech is used, which corresponds to the following text: ``A large open fireplace, with rusty dogs in it, and a bare boarded floor; at the far end, fleeces of wool stacked up; in the middle of the floor, some empty corn-bags.''}
\label{overview}
%\vspace{-0.5cm}
\end{figure}

\section{Related Work}

\subsection{Spoken Question Answering}
{
The task of spoken question answering (SQA) involves providing a segment of speech and requiring responses based on the content of the speech and questions. Depending on the type of response, it can be divided into three forms: time span }\cite{spokensquad,odsqa}, {option label} \cite{toefl}{, and free form} \cite{spokencoqa}.
{The methods for predicting the time span are the most numerous. Chuang et al. }\cite{speechbert} {first leveraged a text dataset to pre-train a textual BERT. Subsequently, they trained the initial phonetic-semantic joint embedding and eventually trained a shared BERT model that utilized both text and speech, including pre-training with the Masked Language Model (MLM) task, followed by fine-tuning on the SQA task. Chung et al. }\cite{splat} {separately pre-trained speech and language modules using speech and text data via MLM task, and aligned the representations from both modules by leveraging a small amount of paired speech and text data, then they used the speech module to extract the aligned speech feature for further fine-tuning on SQA task. You et al. }\cite{threetasksqa} {proposed three auxiliary self-supervised tasks for SQA, and then to learn noise-invariant utterance representations in a contrastive objective by adopting multiple augmentation strategies. In addition, they also designed Temporal-Alignment attention to semantically align the speech-text clues. Papers that answer in the form of option labels mainly appeared in the early stage and can generally be addressed through a multi-classification model. Chung et al. }\cite{optionmodel} {explored the beneficial effects of transfer learning on the TOEFL dataset }\cite{toefl}. {The study examined both the end-to-end memory network and the query-based attention CNN. With the MovieQA dataset }\cite{moiveqa} {as the source task, the performance on the TOEFL dataset improved by 7\% compared to previous results. Models capable of answering in free form are primarily multimodal LLMs, which we will introduce in the next subsection.}

%In addition to collecting the Spoken-CoQA dataset, You et al. \cite{spokencoqa} also proposed a teacher-student paradigm with an ASR module to process spoken documents or spoken questions to get the start and end index.
\subsection{End-to-end Speech-based Large Language Model}
%随着文本单模态大模型的成功，多模态大模型也在逐步获得关注。在语音多模态大模型方面，早期的工作通过chatgpt理解用户意图，然后调用符合的模型来解决对应问题。在本文同期的工作中出现了end-to-end的语音多模态大模型，包括可以解决ASR问题，word replacement，summarization等固定任务的3，他们通过confermer encoder处理语音，并通过cross-attention接入LLM。4针对了ASR，语音翻译和SQA，对于ASR和语音翻译，他们使用了固定的instructions，对于SQA则使用了自由的形式，他们的语音通过任意固定的encoder进行特征提取后，通过2层可训练的transformer后拼接入LLM。Pengi针对了21种常见的语音和audio任务,使用了来自CLAP的audio transformer作为audio encoder，然后将特征拼接进入LLM。
{With the success of text-based unimodal LLMs, multimodal LLMs are gradually gaining attention. In concurrent work with this article, there emerged end-to-end speech-based multimodal LLMs. For instance, Lai et al. }\cite{compare1} {targeted fixed tasks like ASR, word replacement, and summarization. They process the speech through a Conformer encoder and integrate it with an LLM via cross-attention. They also used the LibriSpeech dataset }\cite{librispeech} {as their speech source for training. Wang et al. }\cite{compare2} {tackled ASR, speech translation, and SQA. For ASR and speech translation, they employed fixed instructions, while for SQA, a free-form approach was used. Their speech feature, after feature extraction through an arbitrary fixed speech encoder, was fed into a two-layer trainable transformer before being concatenated with the LLM. They utilized the multilingual YouTube dataset }\cite{compare2data1}{, CoVoST2 dataset }\cite{compare2data2}{, and Alpaca dataset }\cite{compare2data3} {as their speech sources for training. Deshmukh et al. }\cite{compare3} {focused on 21 common speech and audio tasks, using a transformer encoder from CLAP }\cite{clap} {as the audio encoder, which then concatenated features with the LLM. They sourced their training datasets from 21 task-related datasets.}

%With the success of text-based unimodal LLMs, multimodal LLMs are gradually gaining attention. Concurrently with this article, there emerged end-to-end speech-based multimodal LLMs. For instance, Lai et al. \cite{compare1} targeted fixed tasks such as ASR, word replacement, and summarization. They processed speech with a Conformer encoder and integrated it into a LLM via cross-attention. Specific to ASR, speech translation, and SQA (Spoken Question Answering), they employed fixed instructions for ASR and speech translation, while adopting a free-form approach for SQA. The speech features were extracted using any specified encoder and then linked to the LLM after processing through two trainable transformer layers. Meanwhile, Pengi targeted 21 common speech and audio tasks using an audio transformer from CLAP as the audio encoder, subsequently integrating these features into the LLM.
\section{The LibriSQA Dataset}

\begin{figure}[ht]
    \centering
    \subfigure{\includegraphics[width=20pc]{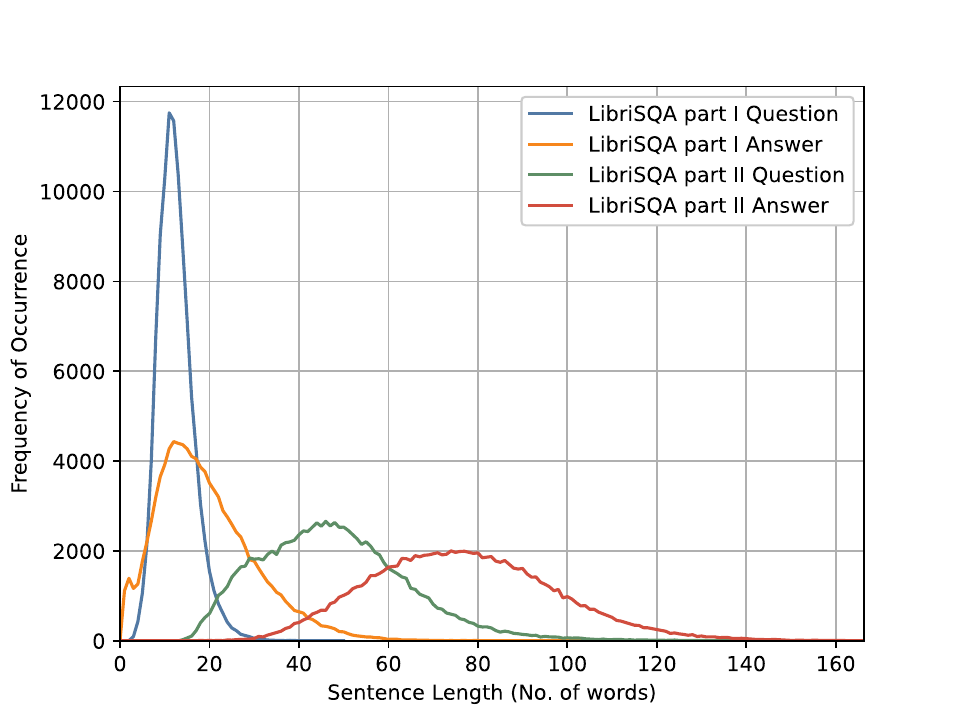}}
\caption{{This figure depicts the word frequencies counted on LibriSQA.} }
\label{dataanalysis0}
\end{figure}

\begin{figure*}[ht]
    \centering
    %\subfigure{\includegraphics[width=15.5pc]{sentence_length.pdf}}
    \subfigure{{\includegraphics[width=21pc]{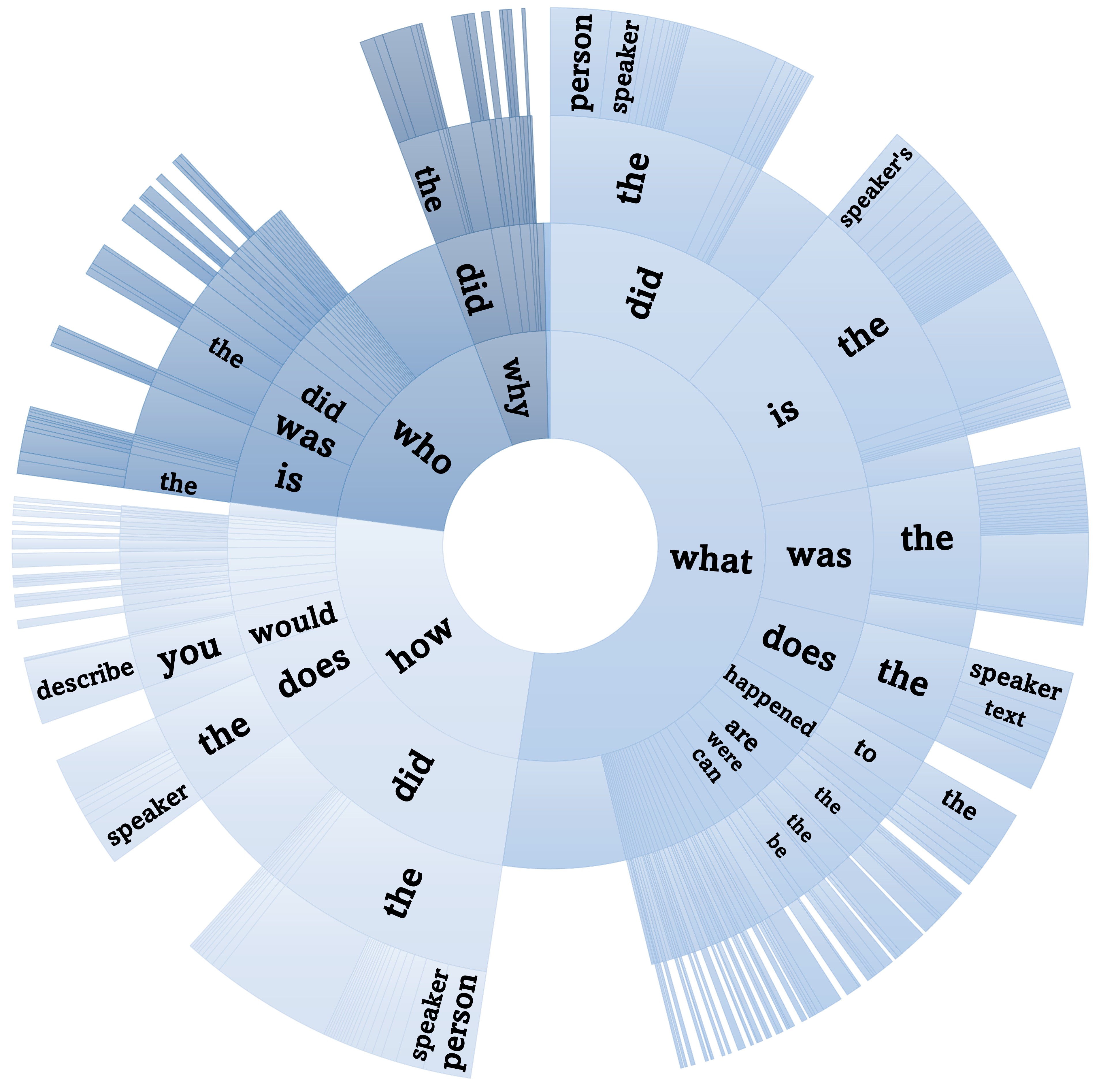}}}
    %\hspace{1em}
    \subfigure{{\includegraphics[width=21pc]{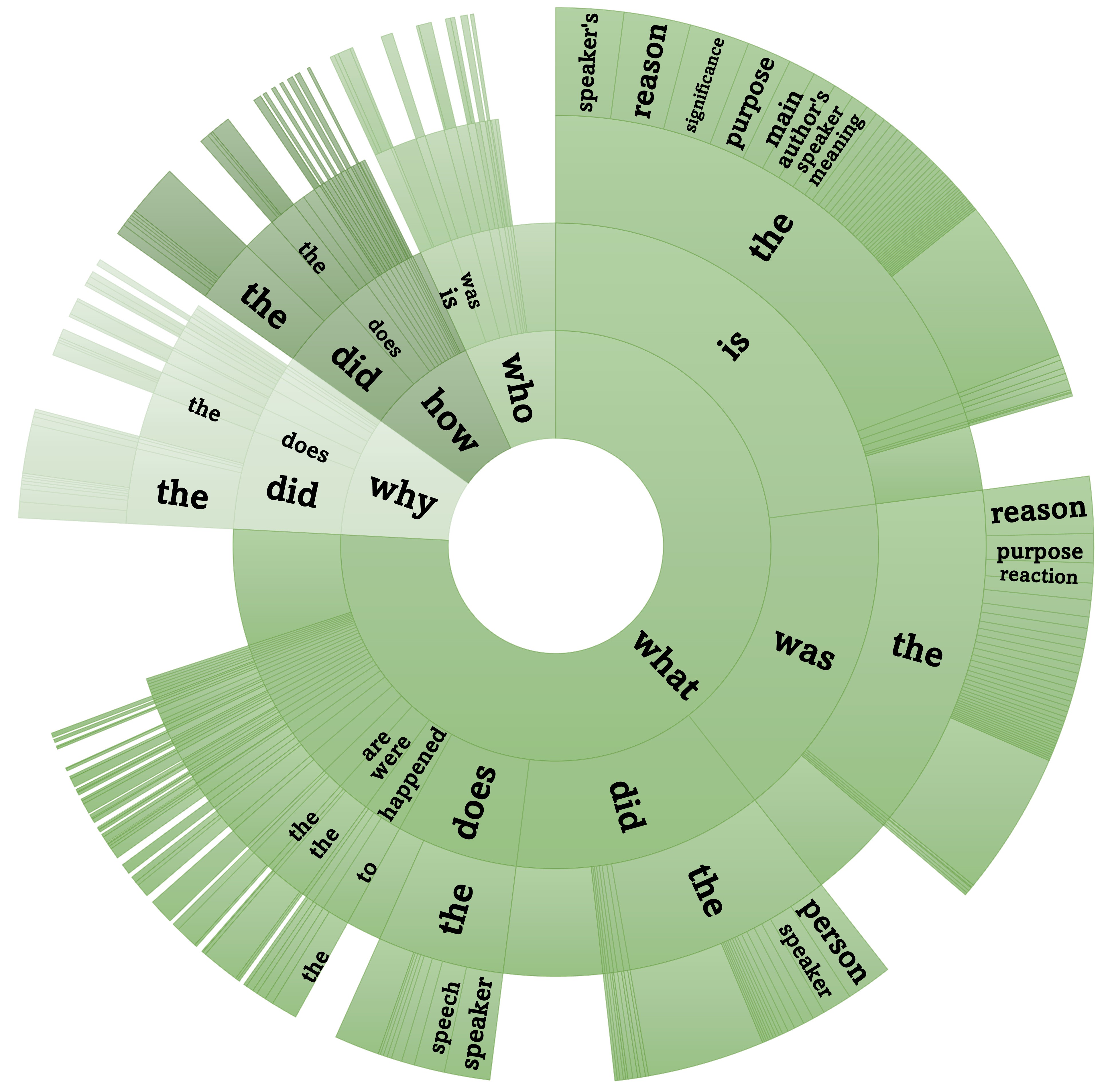}}}
    %\hspace{1em}

\caption{{The two figures, from left to right, represent the distribution of questions for LibriSQA Part I and Part II, respectively. The larger the space occupied by a word, the higher its frequency of occurrence. Due to space limitations, some samples are not included in the statistics.}}
    \label{dataanalysis1}
\end{figure*}

{We believe that an ideal SQA dataset designed for LLMs should have the following attributes:

1) It should accommodate free-form, open-ended QA. The answer should require reasoning instead of being a simple extraction from the spoken document.

2) Evaluating the model's performance on the dataset should be straightforward and efficient.

3) The duration of the speech segments should be concise to seamlessly integrate with LLMs without the need for compression.

4) To enhance real-world applicability, the speech samples should ideally be authentic rather than synthetic.}

To achieve these goals, we introduce the LibriSQA dataset, which is collected with the help of ChatGPT. LibriSQA is divided into two parts, each consisting of 107k question-answer pairs. The difference between LibriSQA Part I and Part II lies in their format. While each question in Part II has four options, the corresponding answer specifies the correct choice and provides an analytical rationale. Part I presents questions without options, and the answers are complete sentences. A representative sample of the LibriSQA dataset is shown in Figure~\ref{overview}, specifically in the third and fourth scenarios. We believe that each part of the dataset has unique characteristics that justify their retention. Part I has a more natural structure, while Part II is more suitable for straightforward evaluation, with accuracy as a viable criterion. A comparison of different LibriSQA and other representative SQA datasets is also summarized in Table~\ref{datasetcompare}.

LibriSQA is developed based on LibriSpeech \cite{librispeech}. LibriSpeech is a comprehensive dataset that consists of approximately 1000 hours of 16kHz English read speech. It has been carefully selected from audiobooks within the LibriVox project \cite{librivox} and subsequently segmented and aligned. To construct our training set, we have utilized the ``360h clean'' speech subset from LibriSpeech. Our reasons for developing the SQA dataset using LibriSpeech are threefold: 1) LibriSpeech provides data specifically relevant to Automatic Speech Recognition (ASR) tasks, which can serve as an auxiliary to the SQA task. We will verify this proposition in subsequent experimental sections. 2) The recordings in LibriSpeech are from authentic human narrators, rather than synthesized Text-To-Speech (TTS) counterparts. This ensures a more accurate reflection of real-world scenarios. Additionally, the large amount of data in LibriSpeech provides a sufficient sample size. 3) Most speech samples in LibriSpeech are concise, typically staying within 20 seconds. This contrasts with existing SQA datasets, making LibriSQA more suitable for integration into LLMs, especially for memory-intensive deployments on standard computing devices.

%We have structured LibriSQA into two distinct parts: Part I contains questions and their direct answers for each sample. In contrast, Part II adopts a multiple-option format, with answers providing both the correct selection and an accompanying reasoning. Part I offers a raw, natural question-answer interface, though its inherent flexibility poses challenges during evaluation, given the myriad plausible answers to a single question. In contrast, Part II's structured format facilitates straightforward accuracy-based evaluations, even if the multiple-option paradigm is not reflective of natural conversational dynamics. We choose to retain both parts, catering to diverse application scenarios.
We have meticulously organized LibriSQA into two clearly defined sections: Part I presents a raw, unadulterated question-answer interface, which, while incredibly adaptable, presents evaluation challenges due to the multitude of plausible answers to a single question. Conversely, Part II's structured format permits straightforward accuracy-based evaluations, albeit at the expense of reflecting natural conversational dynamics through a multiple-choice paradigm. Our decision to retain both parts stems from our commitment to meeting the needs of various application scenarios.

%Moreover, considering that the LibriSpeech dataset is traditionally deployed for ASR tasks, we postulate its potential morphing into an SQA-like framework. For instance, presenting the model with a speech segment and posing the question, 'Transcribe this speech segment into text'. To test this hypothesis, we generated a hundred descriptive prompts tailored for the ASR task using ChatGPT, subsequently designating a random description as a query for each sample.

\subsection{Question-Answer Pair Generation of LibriSQA Part I/II}
In order to generate QA pairs efficiently and with high quality, we use ChatGPT because of its powerful capability in text generation. We input the textual document of each speech segment into ChatGPT, and for Part I we use the following prompt: 

\emph{HERE IS THE TEXTUAL DOCUMENT FROM LIBRISPEECH. Generate a question-answer pair based on the above information only. Do not use any other knowledge outside of this text. The question and answer need to be complete sentences.}

For Part II we use the following prompt: 

\emph{
HERE IS THE TEXTUAL DOCUMENT FROM LIBRISPEECH.
Generate a question along with 4 options and give a corresponding answer and analysis based on the above information only. Do not use any other knowledge outside of this text.
Your answer needs to use the following format, and the parts in $<>$ are what you need to fill in:
Question: $<>$
Options:
A. $<>$
B. $<>$
C. $<>$
D. $<>$
Answer and analysis: The answer is Option $<>$ because Option A $<>$, Option B $<>$, Option C $<>$, Option D $<>$
}

For each speech segment, we generate a singular QA pair. This decision results from the inherent brevity of individual speech segments in LibriSpeech. Should multiple QA pairs be derived from the same speech segment, there is a significant likelihood of reduced diversity, ultimately leading to similar questions and answers. Given the total duration of the speech content, producing only one QA pair for each segment already ensures a sufficient sample size within our dataset. Utilizing ChatGPT, we get 107k pairs for both Part I and Part II. 
\subsection{Data Processing of LibriSQA Part I/II}

%The data obtained from Chatgpt needs to be filtered, this is because Chatgpt's answers do not follow the prompts completely. For Part I, 1) Whether the QA pair is relevant to the text. So we input the QA pair and text into chatgpt and let it score from 0 to 10 to evaluate the relevance. A random selection of the lowest scoring samples is manually checked and it is concluded that even the lowest scoring QA pairs are relevant to the text. 2) Whether the question can be answered using the given text alone without using external knowledge. To solve this problem we let Chatgpt determine yes or no, and for samples that are answered no the generation is repeated until the answer is yes. 3) the most common problem is that for some short texts, chatgpt repeats the text directly in the question, which causes the SQA question to degenerate into a textual QA question. This problem can be easily solved with some string processing methods. 4) We simply replace the word "text" with the word "speech" for all the words referring to the background text in the QA pair, because in the dataset we actually use speech rather than text for SQA. For Part II, the first four steps are the same as for Part I, except that Part II has the additional step of confirming the number of options and whether there are any duplicates.

{Data derived from ChatGPT needs rigorous processing, primarily due to occasional deviations of ChatGPT's responses from the provided prompts. For Part I, our filtration process includes the following criteria:}

\textbf{Relevance to the document}: To ascertain the relevance of each QA pair to its associated textual document, we feed the QA pair and the textual document into GPT-4 for scoring on a 0-10 scale. A manual inspection of a random subset of low-scoring samples confirms that even these pairs remain contextually relevant. Therefore, we do not process them any further.

\textbf{Document sufficiency for answering}: The objective is to ensure that the questions can be satisfactorily answered using only the given textual document, without any external knowledge requirements. To facilitate this, we leverage GPT-4 to assess the sufficiency of the textual document in providing answers, with iterative regeneration for those considered insufficient.

\textbf{Direct repetitions in short texts}: A common issue observed is ChatGPT's tendency to directly repeat the given textual document within the posed question, unintentionally transforming the task into a more fundamental textual question-answering task. We rectify this problem through simple string processing techniques.

\textbf{Terminology refinement}: A common sentence in the QA pairs is ``According to the text...''. Given our dataset's real document is in spoken form rather than text, all references to ``text'' within the QA pairs are simply replaced with ``speech''.

For Part II, the initial methodology is similar to the four steps for Part I. However, Part II introduces an additional criterion to ensure there are no duplicate options and that the number of options is neither excessive nor insufficient.

\subsection{Data Analysis of LibriSQA Part I/II}
%Figure \ref{dataanalysis} shows the results of the data analysis performed for LibriSQA. Based on the chart on the left, we can see that Part I questions and answers are all shorter, while Part II questions and answers are longer, which is to be expected as Part II contains additional options and analysis sections. By comparing the graph in the center to the graph on the right, it is evident that for both Part I and Part II of LibriSQA, questions predominantly commence with "what". However, the proportion of such questions is significantly greater in Part II. We hypothesize that this can be attributed to Part II consisting of questions with multiple choices. Questions beginning with "What" often seek specific information, which can more straightforwardly facilitate the formulation of structured options, in contrast to the more open-ended questions typified by prompts like "Why" or "How", which are more prevalent in Part I. Beyond that, there are no words that clearly predominate after the first word.
%As for the distribution of correct choices in Part II, choice A accounts for 21.7\%, choice B for 30.3\%, choice C for 32.2\% and choice D for 15.8\%.

Figure~\ref{dataanalysis0} and Figure~\ref{dataanalysis1} present a comprehensive data analysis of LibriSQA. Figure~\ref{dataanalysis0} shows a noticeable frequency discrepancy between Part I and Part II questions and answers: the former tends to be more concise. This aligns with our expectations, as Part II includes additional option sentences and analytical components.

%The central and right graphs reveal a prevalent initiation of questions with "what" in both Part I and Part II of LibriSQA. Intriguingly, the prevalence is notably higher in Part II. We suppose this phenomenon is due to the nature of multiple-option questions in Part II. Questions beginning with "What" often seek specific information, which can more straightforwardly facilitate the formulation of structured options, in contrast to the more open-ended questions typified by prompts like "Why" or "How", which are more prevalent in Part I. Beyond the first word, no specific term significantly dominates.

Figure~\ref{dataanalysis1} indicates a significant prevalence of questions starting with ``what'' in both Part I and Part II of LibriSQA. Interestingly, this prevalence is even higher in Part II. We believe this phenomenon is influenced by the nature of multiple-choice questions in Part II. ``What'' questions often seek specific information, making it easier to create structured options compared to the more open-ended questions that are common in Part I, such as prompts starting with ``why'' or ``how''. Apart from the first word, no other specific term dominates significantly.

As for the distribution of correct answers in Part II, the distribution is shown below: choice A represents 21.7\%, choice B encompasses 30.3\%, choice C stands at 32.2\%, and choice D is at 15.8\%.

\subsection{Reformulation of LibriSpeech}
\label{reform}
%Considering the initial designation of the LibriSpeech dataset for ASR tasks, wherein it provides the corresponding text for each speech sample, it holds the potential to facilitate the alignment of speech and text modalities, potentially enhancing the model performance in SQA tasks. Therefore, we intend to integrate it into our dataset. However, it is worth noting that LibriSpeech does not originally follow the SQA format as it lacks textual questions. To address this, we leverage ChatGPT to create one hundred prompts for ASR tasks, such as "Transcribe this speech segment into the text". Each sample is then randomly paired with a prompt to serve as its question. Finally, we complete the reformulation of LibriSpeech by using the speech as the spoken document and the corresponding text as answers.

{Considering the initial designation of the LibriSpeech dataset for ASR tasks, which provides the corresponding text for each speech sample, it has the potential to facilitate the alignment of speech and text modalities. This alignment can potentially enhance the model performance in SQA tasks. Therefore, we intend to utilize the original label of LibriSpeech.
However, it is important to note that LibriSpeech does not originally follow the SQA format as it lacks textual questions. To address this, we leverage ChatGPT to create one hundred prompts for ASR tasks, such as ``Transcribe this speech segment into text''. Each sample is then randomly paired with a prompt to serve as its question as the complete form of SQA requires questions, spoken documents, and answers. Finally, we complete the reformulation of LibriSpeech by using the speech as the spoken document and the corresponding text as the answers. Please note that the reformed data discussed in this part are not included in LibriSQA Part I/II.}

\begin{figure}[ht]
\centerline{\includegraphics[width=22pc]{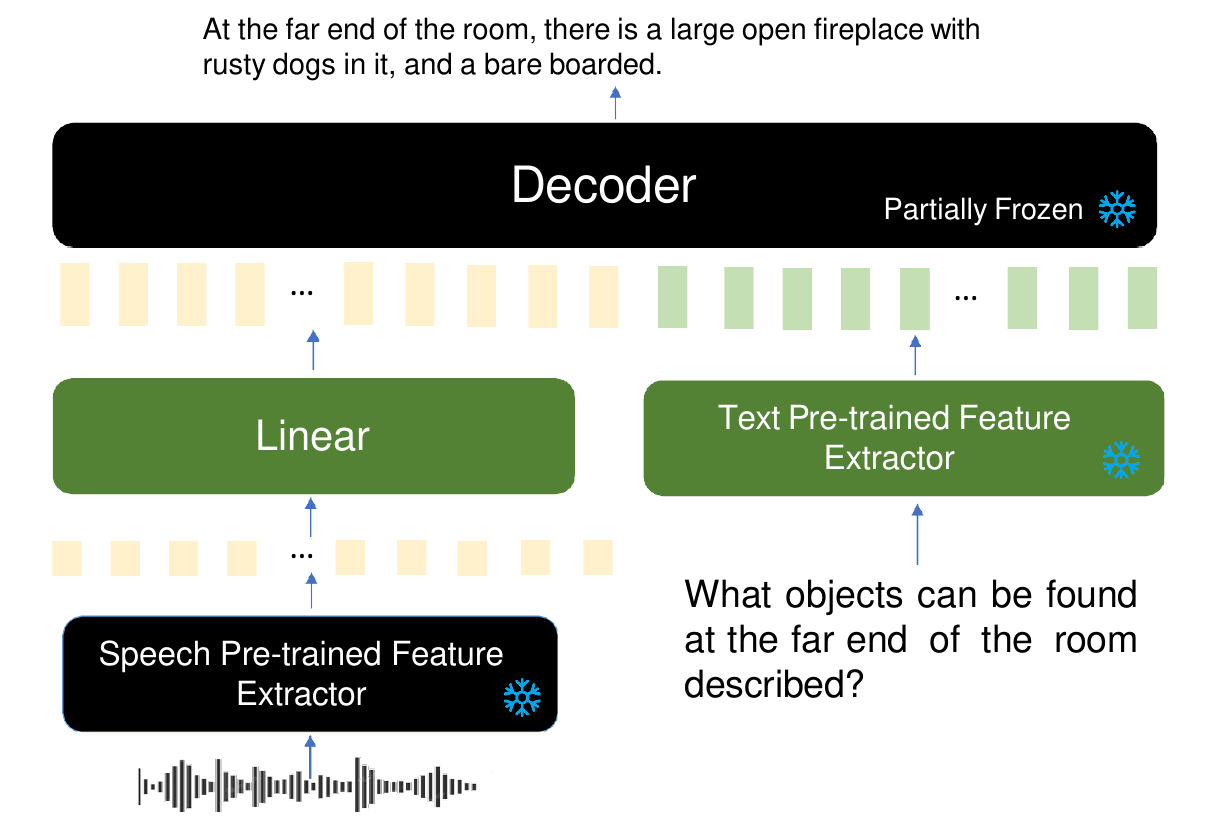}}
\caption{The architecture of our framework. The speech shown here corresponds to the text of ``A large open fireplace, with rusty dogs in it, and a bare boarded floor; at the far end, fleeces of wool stacked up; in the middle of the floor, some empty corn-bags.''}
\label{framework}
\end{figure}

\section{Method}

%To accomplish the SQA task, we design a lightweight and end-to-end framework as shown in Fig \ref{framework}. In this section, we will first describe the SQA setting we use in detail, then introduce the two important parts which are the feature extraction module and decoder module respectively, and finally describe the training process of the model.
As mentioned in Section \ref{intro}, traditional SQA methodologies primarily focus on predicting time frames, which are not suitable for adaptation to LibriSQA. Therefore, in this paper, we also propose a novel framework specifically designed to handle the complexity of this challenging dataset. To ensure that LLMs can truly understand speech content and demonstrate their independent ability to do so, we train our model in an end-to-end manner, without relying on the ASR module for speech content processing.
This lightweight and end-to-end framework is depicted in Figure~\ref{framework}. In the following discussion, we will start by explaining the specific SQA setting used. Then, we will describe the two main components: the feature extraction module and the decoder module. Finally, we will provide a detailed explanation of the model's training methodology to conclude this section.

\subsection{Problem Formulation}
%Given a dataset $\chi  \in \left\{ {{S_i},{Q_i},{A_i}} \right\}_i^N$, SQA task requires training a model that output ${\tilde A_i} = {\rm{Model}}\left( {{S_i},{Q_i};W} \right)$, where $S_i$ is the speech that provides background information, $Q_i$ is the question that in text or speech modality, $A_i$ is the answer that in text modality and ${\tilde A_i}$ is the output from the model. Following the convention by Spoken-SQuAD \cite{spokensquad}, we opt for a textual modality for the question. We posit that this setting is more challenging, as it necessitates that the model master bimodal alignment between speech and text concurrently. The model's output is typically presented in one of two formats. The first, a prevalent approach, entails predicting a specific time span wherein the answer consists of words directly derived from the background information. The second, a more natural approach, requires the model to generate a comprehensive answer sentence. This format offers a higher degree of freedom and closely mirrors real-life scenarios, which is the format we employ. Normally the SQA is set up with $S_i$ containing multiple sentences and $Q_i$ and $A_i$ consisting of single sentences. LibriSQA Part II is an exception because it comes with options and analysis, so $Q_i$ and $A_i$ also contain multiple sentences.

Given a dataset $\chi \in \left\{ {{S_i},{Q_i},{A_i}} \right\}_i^N$, the SQA task involves training a model that outputs ${\tilde A_i} = {\rm{Model}}\left( {{S_i},{Q_i};W} \right)$. Here, $S_i$ represents the speech providing background information, $Q_i$ represents the question in text or speech form, $A_i$ represents the answer in text form, and ${\tilde A_i}$ represents the model's output. Following the convention established by Spoken-SQuAD \cite{spokensquad}, we choose to use textual modality for the question. We believe this setting is more challenging as it requires the model to simultaneously understand and align speech and text.
The model's output is typically presented in one of two formats. The first format involves predicting a specific time span where the answer consists of words directly derived from the background information. The second format, which we employ, requires the model to generate a comprehensive answer sentence. This format offers more freedom and closely resembles real-life scenarios. In most SQA setups, $S_i$ consists of multiple sentences, while $Q_i$ and $A_i$ consist of single sentences. However, LibriSQA Part II is an exception, as it includes options and analysis, resulting in multiple sentences in both $Q_i$ and $A_i$.

\subsection{Feature Extractor}
\label{extractor}
{In this paper, pre-trained speech models are utilized for feature extraction given their capabilities of learning rich speech representation, which include} wav2vec 2.0 \cite{wav2vec2}, HuBERT \cite{hubert} and WavLM \cite{wavlm}. {To avoid the potential discrepancies in feature dimensions provided by different pre-trained models, we subsequently incorporate linear layers to facilitate feature mapping. Furthermore, we also investigate a rather simple strategy that leverages solely this linear layer, targeting the alignment of speech and text modalities. Throughout the training phase, the parameters of the pre-trained speech models remain frozen, with only the linear layer engaging in the training procedure. As for the text modality, we directly employ the embedding layer from LLMs, which similarly remains frozen during the training process. In the subsequent stages, we concatenate the speech features with the text features, forwarding them to the decoder for further processing.}

{These choices of speech feature extractors are not absolute and they can be replaced with other speech feature extractors. It's important to note that these models may have both pre-trained versions and versions fine-tuned for ASR tasks. The ASR fine-tuned versions are likely to have utilized annotations from the LibriSpeech dataset, which could lead to the leakage of textual information, essentially degrading the SQA task into a mere text-based QA task. To circumvent this issue, the speech feature extractors employed in this paper are the pre-trained versions of each model. In this way even if the LibriSpeech data was involved in the pre-training, the lack of use of its annotations prevents the introduction of bias. Moreover, if unsupervised training of the speech data on speech feature extractors proves beneficial for this SQA task, we can apply unsupervised training regardless of the speech data on which LibriSQA is based. To further investigate whether the LibriSpeech fine-tuned versions of the speech feature extractor could lead to textual information leakage, we will provide the results of the experiment in }section~\ref{ftextractor}.

{Here are some brief descriptions of the speech feature extractors we use:}

\textbf{Wav2vec 2.0} \cite{wav2vec2}. {Wav2vec 2.0 is a pre-trained speech model for self-supervised learning of speech representations which masks latent representations of the raw waveform and solves a contrastive task over quantized speech
representations. It is the first to show that learning powerful representations from speech audio alone followed by fine-tuning on transcribed speech can outperform the best semi-supervised methods.}

\textbf{HuBERT} \cite{hubert}. {HuBERT is another pre-trained model for self-supervised learning of speech model that relies on predicting K-means cluster assignments of masked segments of continuous input. Compared to wav2vec 2.0, HuBERT adopts a more direct predictive loss by separating the acoustic unit discovery step from the masked prediction representation learning phase rather than using a contrastive loss that requires careful design.}

\textbf{WavLM} \cite{wavlm}. {WavLM model aims to extract universal speech representations to solve both ASR and non-ASR tasks while wav2vec 2.0 and HuBERT mainly aim to solve ASR tasks. WavLM extends the HuBERT framework to masked speech prediction and denoising modeling and also employs gated relative position bias for the Transformer structure to better capture the sequence ordering of input speech.}
%For the extraction of speech features, numerous pre-trained models are currently available to fulfill this task. Therefore, we endeavor to employ these pre-trained acoustic feature extractors, such as Wav2vec2, wavlm, and hubert. Due to potential disparities in feature dimensions extracted by various pre-trained models and the decoder, we subsequently employ linear layers to perform mapping. Moreover, we also explore a simpler approach involving the use of this linear layer exclusively, aiming to achieve alignment between the speech and text modalities. The parameters of the speech pre-trained models remain entirely frozen during training, with only the linear layer engaging in the training procedure. For the text modality, we simply use the embedding layer from LLMs and it's also frozen during training. Finally we concatenate the speech features with the text features and pass them to the decoder in the next step.

%For the extraction of speech features, numerous pre-trained models can satisfy this need. 

\subsection{Decoder}
We use the open-source model LLaMA \cite{llama} as our decoder. Like most LLMs, LLaMA used the Transformer structure \cite{transformer}, with the main difference being the use of pre-normalization \cite{prenorm}, SwiGLU activation function \cite{glu}, and Rotary Embeddings \cite{rot}. With different numbers of parameters, four versions of LLaMA models have been released: 7B, 13B, 33B, and 65B. LLaMA-13B outperformed GPT-3 (175B) on most benchmarks while LLaMA-65B was competitive with the best models \cite{llama}. In this work, we use the 7B version as our pre-trained language decoder. 

%To make our framework more lightweight and efficient, we use LLaMA-adapter \cite{llamaadapter} to efficiently fine-tune LLaMA. When using llama-adapter, the parameters of LLaMA itself are frozen, and additional learnable adaption prompts are added for fine-tuning. Because the prompts are randomly initialized, they may introduce perturbations during the initial phase of training to affect the training, so zero-initialized attention is designed to solve this problem. They also propose to fuse multimodal information by adding it to the prompt. However, we argue that this is not feasible for speech, especially for the SQA task, because without compression of the speech information, speech of tens of seconds would require at least thousands of prompts per layer while they only use ten for the visual-question answering (VQA) task, which would result in an even larger number of fine-tuning parameters than the full fine-tuning, thus making efficient fine-tuning impossible. Since in the SQA task, the speech contains the information necessary to answer the question, we do not think that the lossy compression of the speech is justified. The problem of how to compress it without affecting SQA tasks is something we see as a future work. The way we incorporate speech information is by concatenating speech features with text features and feeding them to the decoder, as previously described, while retaining the learnable adaption prompts and zero-initialized attention techniques in the LLaMA-adapter for fine-tuning.

To improve the lightweight and efficient nature of our framework, we utilize the LLaMA-adapter \cite{llamaadapter} for efficient fine-tuning of LLaMA. With LLaMA-adapter, the parameters of LLaMA itself are frozen, and additional learnable adaptation prompts are introduced for fine-tuning. To address the potential perturbations caused by the randomly initialized prompts during initial training, zero-initialized attention is designed \cite{llamaadapter}.
It also suggests fusing multimodal information by adding it to the adaptation prompts \cite{llamaadapter}. However, we argue that this approach is not feasible for speech, especially in the context of the SQA task. Without compressing the speech information, utilizing speech of long duration would require thousands of prompts per layer, whereas they only use ten for the Visual Question Answering (VQA) task \cite{llamaadapter}. This would result in an excessive number of fine-tuning parameters, making efficient fine-tuning impractical. In the SQA task, the speech contains essential information for answering the question, and we believe that lossy compression of speech is not justified. Exploring methods for compressing speech without affecting SQA tasks is an area we will focus on in future work.
In this work, we incorporate speech information by directly concatenating speech features with text features and feeding them to the decoder, as described earlier. We retain the learnable adaptation prompts and zero-initialized attention techniques in the LLaMA-adapter for fine-tuning.

% \usepackage{multirow}
\begin{table*}[]
\centering % 居中表格
 \caption{Results of the experiments on the ASR task. LibriSpeech-test-clean is used as the test set. The first section shows the results of our framework, the second shows the benefit of SQA for ASR and the third shows the the results of existing methods. ``Partially'' means the speech pre-training is used in the speech feature extractor. ``LS'' is for LibriSpeech. ``NLL'' is for negative log-likelihood. ``CTC'' is for connectionist temporal classification }
  \label{asr}
\begin{tabular}{lcccc|c}
\toprule
\textbf{Methods}                                & \multicolumn{1}{l}{\textbf{Trainable Param}} & \textbf{Speech Pre-training} & \textbf{Labeled Corpus}           & \textbf{Loss}        & \textbf{WER$\downarrow$} \\ \midrule
wav2vec 2.0 Base+LLaMA 7B                       & 4.4M                                   & \multirow{6}{*}{Partially}   & \multirow{6}{*}{Reformed LS 360h} & \multirow{6}{*}{NLL} & 7.96           \\
wav2vec 2.0 Large+LLaMA 7B                      & 5.4M                                    &                              &                                   &                      & 11.34          \\
HuBERT Base+LLaMA 7B                            & 4.4M                                   &                              &                                   &                      & 9.01           \\
\textbf{HuBERT Large+LLaMA 7B} & \textbf{5.4M}                                    &                              &                                   &                      & \textbf{3.95}           \\
WavLM Base+LLaMA 7B                             & 4.4M                                   &                              &                                   &                      & 4.41           \\
WavLM Large+LLaMA 7B                            & 5.4M                                    &                              &                                   &                      & 4.82           \\ \midrule
wav2vec 2.0 Base+LLaMA 7B                       & 4.4M                                   & \multirow{6}{*}{Partially}   & \multirow{6}{*}{LibriSQA w/o ASR dataset}  & \multirow{6}{*}{NLL} & 57.37          \\
wav2vec 2.0 Large+LLaMA 7B                      & 5.4M                                    &                              &                                   &                      & 59.89          \\
\textbf{HuBERT Base+LLaMA 7B}                            & \textbf{4.4M}                                   &                              &                                   &                      & \textbf{27.31}          \\
HuBERT Large+LLaMA 7B                           & 5.4M                                    &                              &                                   &                      & 40.40          \\
WavLM Base+LLaMA 7B                             & 4.4M                                   &                              &                                   &                      & 37.90          \\
WavLM Large+LLaMA 7B                            & 5.4M                                    &                              &                                   &                      & 30.49          \\ \midrule 
wav2vec 2.0 Base \cite{wav2vec2}                                & 95M                                     & Yes                          & LS 100h                           & CTC                  & 6.1            \\
wav2vec 2.0 Large \cite{wav2vec2}                              & 317M                                    & Yes                          & LS 960h                           & CTC                  & 2.7            \\
HuBERT Base \cite{hubert}                                   & 95M                                     & Yes                          & LS 100h                           & CTC                  & 3.4            \\
HuBERT Large \cite{hubert}                                   & 317M                                    & Yes                          & LS 960h                           & CTC                  & 2.0            \\
WavLM Base \cite{wavlm}                                      & 95M                                     & Yes                          & LS 100h                           & CTC                  & 4.6            \\
\textbf{WavLM Large} \cite{wavlm}                                    & \textbf{317M}                                    & \textbf{Yes}                          & \textbf{LS 960h}                           & \textbf{CTC}                  & \textbf{1.8}            \\
Whisper Tiny \cite{whisper}                                   & 39M                                     & Yes                          & 680000h from Internet             & CTC                  & 7.6            \\
Whisper Large \cite{whisper}                                  & 1550M                                   & Yes                          & 680000h from Internet             & CTC                  & 2.7            \\
Instruction-Trained LAS \cite{compare1}                        & 224M                                    & Partially                    & Reformed LS 960h                           & NLL                  & 2.6            \\
SLM \cite{compare2}                                            & 156M                                    & Partially                    & YouTube Corpus                    & NLL                  & 3.2            \\ \bottomrule
\end{tabular}
\end{table*}

\subsection{Training}
In our study, experiments are conducted using the reformed LibriSpeech datasets, specifically LibriSQA Part I and Part II. Since the ASR task has been transformed into the SQA format, we deviate from the conventional use of the Connectionist Temporal Classification (CTC) loss \cite{ctc}, which is typically used in ASR tasks. Instead, we employ the standard negative log-likelihood loss for all three tasks:

%\begin{equation}
%{\cal L}\left( {W|D} \right) =  - \sum\limits_{i = 1}^N {\sum\limits_{j = {p_i} + 1}^{{x_i}} {\log P\left( {{a_{i,j}}|{S_i},{a_{i, < j}};W} \right)} } 
%\end{equation}

\begin{equation*}
{\cal L}\left( {W|D} \right) =  - \sum\limits_{i = 1}^N {\sum\limits_{j = {p_i} + 1}^{{x_i}} {\log P\left( {{a_{i,j}}|{S_i},{a_{i, < j}};W} \right)} } 
\end{equation*}

%The implementation of this particular loss function is intended to enable the model to predict the subsequent word by leveraging both the speech and the current sentence context. In this equation, \(W\) represents the trainable parameters of the model, \(D\) denotes the datasets under consideration, \(N\) signifies the total number of samples within the datasets, \(p_i\) indicates the token count of the prompt in the \(i^{th}\) sample, \(x_i\) stands for the aggregate number of tokens in the \(i^{th}\) sample, and \(S_i\) refers to the speech segment for the \(i^{th}\) sample.
The purpose of implementing this loss is to enable the model to predict the next word by using both the speech and the current sentence context. In this equation, \(W\) represents the trainable parameters of the model, \(D\) represents the datasets being considered, \(N\) indicates the total number of samples in the datasets, \(p_i\) represents the token count of the prompt in the \(i^{th}\) sample, \(x_i\) represents the total number of tokens in the \(i^{th}\) sample, and \(S_i\) refers to the speech segment for the \(i^{th}\) sample.

\begin{figure*}[ht]
    \centering
    \subfigure{\includegraphics[width=21pc]{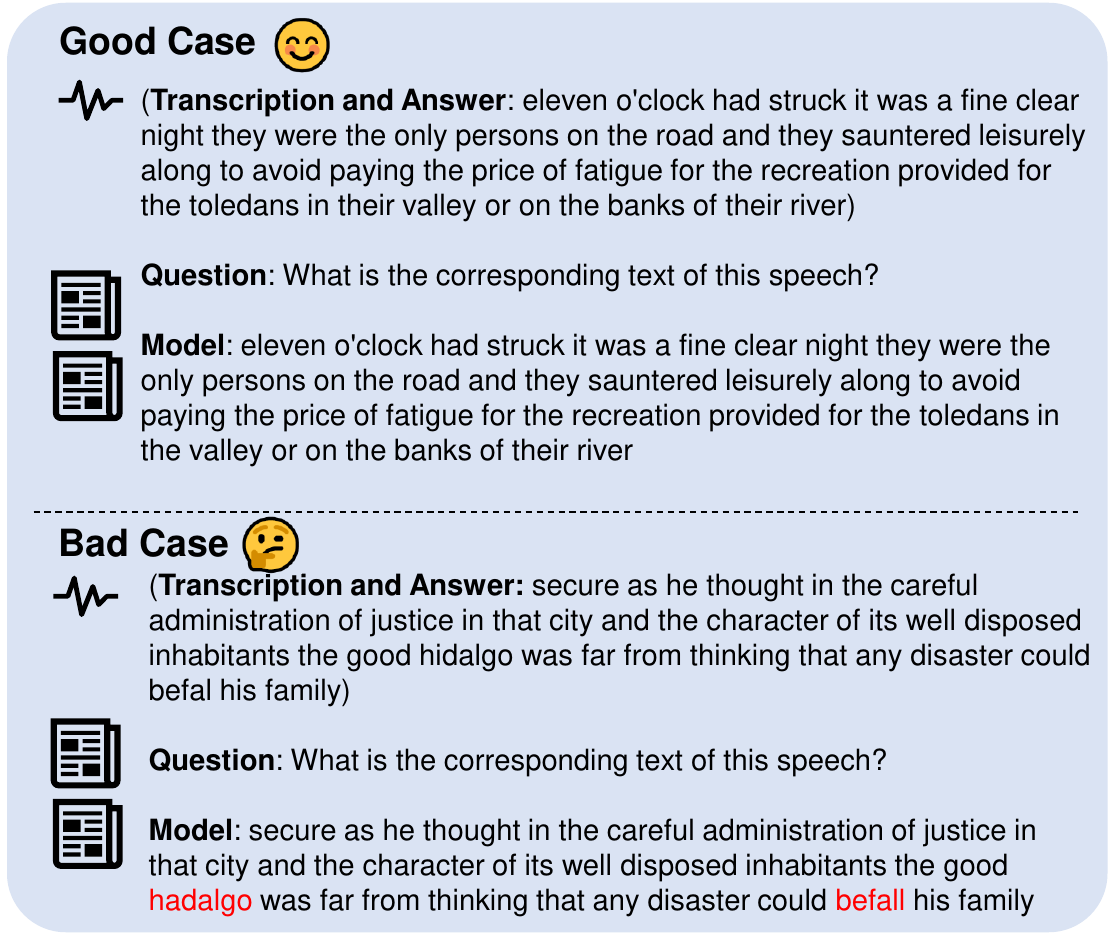}}
    \subfigure{\includegraphics[width=21pc]{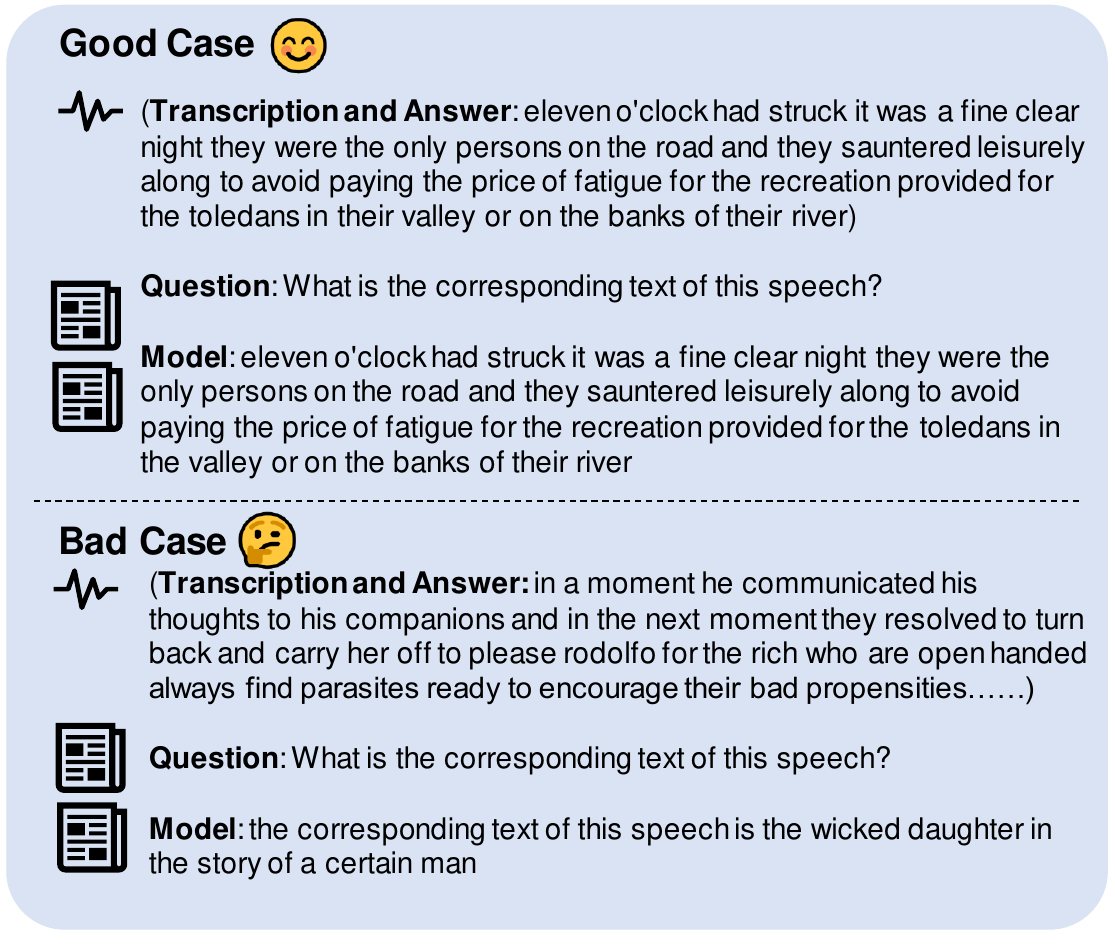}}
\caption{The left figure shows some examples of the ASR task using the model trained with the reformed SQA form of LibriSpeech. The right figure shows some examples using the model trained without using any ASR dataset, and only LibriSQA is used for training.}
    \label{demo01}
\end{figure*}
\section{Experiments}
% \usepackage{multirow}

%In this section, we will initiate a comprehensive discourse on the configurations and evaluation metrics employed within our experimental framework. This will be succeeded by an in-depth exploration of three pivotal components of the experimental design. Initially, we will assess the efficacy of our framework on the Automatic Speech Recognition (ASR) task utilizing the spoken Question-Answering (SQA) format within the LibriSpeech dataset, an elucidation of which is furnished in Section II. Subsequently, we will scrutinize the performance of our model on the LibriSQA Part I subset to underscore its aptitude in natural question-answering scenarios. Furthermore, the model's discerning and analytical capacity will be showcased through its evaluation on the LibriSQA Part II subset.

{In this section, we will begin with a comprehensive discussion of the configuration and evaluation metrics used in our experimental framework. This will be succeeded by an in-depth exploration of pivotal components of the experimental design. First, we will evaluate the effectiveness of our framework on the ASR task using the LibriSpeech dataset in the SQA format which has been described in} Section~\ref{reform}. {Next, we will analyze the performance of our model on the LibriSQA Part I subset, highlighting its ability in natural question answering scenarios. Then, we will demonstrate the model's discerning and analytical capabilities by evaluating it on the LibriSQA Part II subset. Then we will talk about the difference between our end-to-end models and the cascade models. Next, we will discuss the impact of the LibriSpeech fine-tuned speech feature extractors on the SQA task. Finally, we will talk about the computation cost.}

\subsection{Settings and Metrics}
The model we use is LLaMA-7B. Regarding the parameters, we utilized the default settings of the LLaMA-adapter \cite{llamaadapter}, with a learning rate of 9e-3. The learnable adaption prompts were configured with 30 layers, each containing 10 units. Omitting the use of a validation set, we made selections for testing based on the 10th epoch, as determined by observing changes in training set loss for each task. A batch size of 8 was employed, and the training procedures were conducted across two 80GB A100 GPUs for each task. Considering the characteristic sampling rate of 20Hz in the speech pre-trained model, the maximum length for speech feature sequences was set at 900, effectively encompassing the majority of the speech in the LibriSpeech dataset. Additionally, the maximum length for text sequences was established as 300.

For the ASR task, we use the Word Error Rate (WER) as the metric which is the most popular metric for this task. For LibriSQA Part I we use BLEU \cite{bleu}, ROUGE \cite{rouge} and BERT similarity score \cite{bertsimilarity}. BLEU and ROUGE are both geared toward surface-level and structural alignment while BLEU focuses more on precision and ROUGE focuses more on recall. BERT similarity is used to compute semantic similarity. For LibriSQA Part II, because of its multiple-choice format, and the fact that the answer section includes analysis, we use macro accuracy and F1 in addition to BLEU, ROUGE, and BERT similarity to measure whether the model-chosen option is correct or not.

% \usepackage{multirow}

\begin{table*}[]
\centering % 居中表格
  \caption{Results of the experiment on LibriSQA Part I. The metrics primarily assess whether the model's analysis is similar to the reference answers in the dataset. ASR pre-training means whether to use parameters of LLaMA obtained from Table~\ref{asr}. The last row uses random output, and the difference between its result and the other rows reflects the effect of our framework}
  \label{part1}
\begin{tabular}{lcc|ccccc}
\toprule
\textbf{Feature Extractor}         & \textbf{Trainable Param} & \textbf{ASR Pre-training} & \textbf{BLEU-1↑} & \textbf{ROUGE-1$\uparrow$} & \textbf{ROUGE-2$\uparrow$} & \textbf{ROUGE-L$\uparrow$} & \textbf{BERT Similarity$\uparrow$} \\ \midrule
\multirow{2}{*}{wav2vec 2.0 base}  & 4.4M                     & Yes                       & 0.3103          & 0.6093           & 0.4553           & 0.5769           & 0.9127                   \\
                                   & 4.4M                     & No                        & 0.2977          & 0.5954           & 0.4355           & 0.5600           & 0.9143                   \\ \midrule
\multirow{2}{*}{HuBERT base}       & 4.4M                     & Yes                       & 0.3117          & 0.6146           & 0.4563           & 0.5807           & 0.9162                   \\
                                   & 4.4M                     & No                        & 0.3083          & 0.6243           & 0.4654           & 0.5898           & 0.9232                   \\ \midrule
\multirow{2}{*}{WavLM base}        & 4.4M                     & Yes                       & 0.3301          & 0.6475           & 0.4908           & 0.6127           & 0.9270                   \\
                                   & 4.4M                     & No                        & 0.3198          & 0.6437           & 0.4857           & 0.6095           & 0.9287                   \\ \midrule
\multirow{2}{*}{wav2vec 2.0 large} & 5.4M                     & Yes                       & 0.2747          & 0.5660           & 0.4061           & 0.5358           & 0.9112                   \\
                                   & 5.4M                     & No                        & 0.2694          & 0.5337           & 0.3705           & 0.5054           & 0.9103                   \\ \midrule
\multirow{2}{*}{HuBERT large}      & 5.4M                     & Yes                       & 0.3302          & \textbf{0.6538}  & \textbf{0.5019}  & \textbf{0.6209}  & 0.9285                   \\
                                   & 5.4M                     & No                        & 0.3250          & 0.6471           & 0.4921           & 0.6148           & \textbf{0.9307}          \\ \midrule
\multirow{2}{*}{WavLM large}       & 5.4M                     & Yes                       & 0.3338 & 0.6478           & 0.4941           & 0.6130           & 0.9249                   \\
                                   & 5.4M                     & No                        & \textbf{0.3378}          & 0.6482           & 0.4980           & 0.6158           & 0.9251                   \\ \midrule
Random                             & 4.4M                     & No                        & 0.0725          & 0.0947           & 0.0053           & 0.0843           & 0.7874                   \\ \bottomrule
\end{tabular}
\end{table*}

\begin{figure}[ht]
\centerline{\includegraphics[width=20pc]{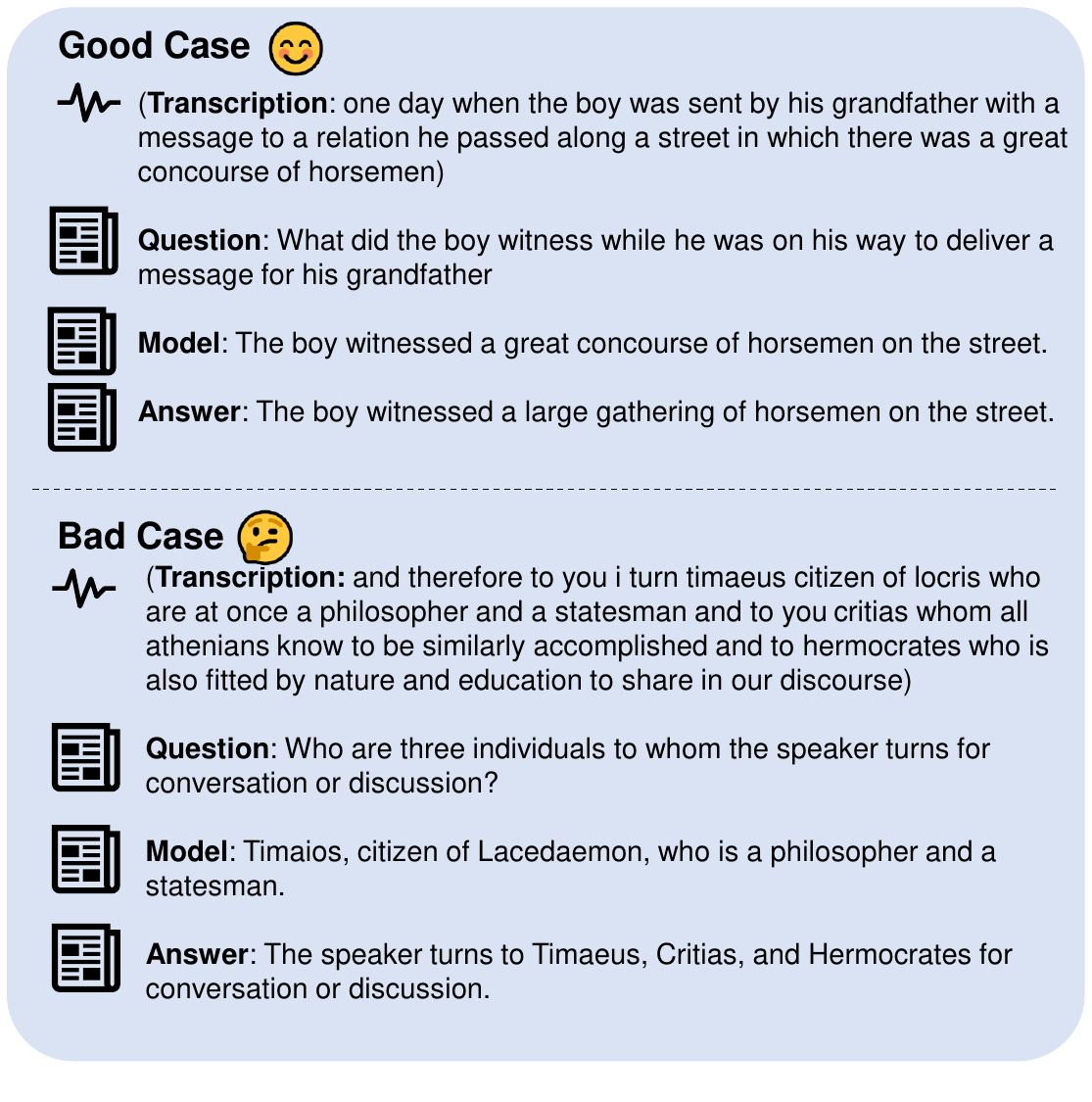}}
\caption{Examples of the model trained with LibriSQA Part I.}
\label{demo2}
\end{figure}

\subsection{The Results of ASR}
\vspace{-0.1cm}
In Table~\ref{asr} {we show the results of the ASR task. Specifically, we use LibriSpeech-test-clean as our test set. LS 100/360/960h are three versions of the LibriSpeech training set, and they differ in total speech length. There are three parts in }Table~\ref{asr}. {The first part, counting from the top, presents the results of models trained using NLL loss on the reformed LibriSpeech 360h dataset discussed in }Section~\ref{reform}. {LibriSQA is not involved in this part. The second part highlights the results of training exclusively on LibriSQA, without the inclusion of any speech-to-text pair datasets such as LibriSpeech, etc. The third part shows the results from the current ASR models. It is noteworthy that the implementations of wav2vec 2.0, HuBERT, and WavLM in the third part are not identical to those in the first and second parts: we omit the language modeling head layers, using them purely for feature extraction. Observing the WER outcomes in the first part, it is evident that our novel training method using NLL loss for ASR tasks is effective, aligning with observations made in }\cite{speechgpt,audiopalm}. {However, as they did not provide WER metrics on LibriSpeech, their results have been omitted from }Table~\ref{asr}. {Additionally, it is obvious that the choice of speech feature extractor significantly impacts the results.}

{The experiments in the second part are conducted to demonstrate the benefits of the SQA task for the ASR task. This part of the experiment is similar to zero-shot inference. Our training does not involve any ASR datasets including the reformed LibriSpeech discussed in }Section~\ref{reform}. {The results indicate that solely training on the SQA dataset can accomplish the ASR task during inference, though the outcomes are considerably influenced compared to the first part, which is expected. An analysis of model outputs reveals that instances with exceptionally high WER often have no relevant ASR textual content. However, for samples where the model genuinely comprehends the prompt and produces ASR text, there is a marked reduction in WER. This suggests that the results are highly influenced by the prompt, and prompt engineering might lead to significant enhancements under this setting.} 

{Comparing the performance of our results in the first Part with the existing ASR model in the third part, our approach achieves similar results to existing models at about one percent of the number of trainable parameters, whether it is a traditional model trained using CTC loss or a large language model similar to ours trained using NLL loss. In addition, our approach helps to integrate the ASR task into a more generalized text generation paradigm than the traditional approach using CTC loss. The results of the second part will not be compared to the third Part because the second part is only a zero-shot inference experiment without using a labeled ASR dataset. Examples can be found in }Figure~\ref{demo01}. {We can see that with LibriSpeech for training, there are few errors even in the bad case (left figure). With only LibriSQA for training, there can be large deviations in the bad case (right figure), but this is to be expected since the training data does not contain any ASR dataset.}

% \usepackage{multirow}

\begin{table*}[]
\centering % 居中表格
\caption{Results of the experiment on LibriSQA Part II. Macro accuracy and F1 score are calculated based on the model's selection among the four options, while other metrics primarily assess whether the model's analysis is similar to the reference answers in the dataset. B means base. L means Large. Param means trainable parameter. Pre-training means whether to use parameters of LLaMA obtained from Table~\ref{asr}}
\label{part2}
\begin{tabular}{lcc|ccccccc}
\toprule
\multicolumn{1}{l}{\textbf{Feature}} & \multicolumn{1}{l}{\textbf{Param}} & \multicolumn{1}{l|}{\textbf{Pre-training}} & \textbf{Accuracy$\uparrow$} & \textbf{F1$\uparrow$}     & \textbf{BLEU-1$\uparrow$} & \textbf{ROUGE-1$\uparrow$} & \textbf{ROUGE-2$\uparrow$} & \textbf{ROUGE-L$\uparrow$} & \textbf{BERT Similarity$\uparrow$} \\ \midrule
\multirow{2}{*}{wav2vec 2.0 B}              & 4.4M                                         & Yes                                            & 0.5846            & 0.5943          & 0.1695          & 0.5934           & 0.3927           & 0.4621           & 0.9167                   \\
                                               & 4.4M                                         & No                                             & 0.5554            & 0.5720          & 0.1683          & 0.5922           & 0.3810           & 0.4583           & 0.9160                   \\ \midrule
\multirow{2}{*}{HuBERT B}                   & 4.4M                                         & Yes                                            & 0.6370            & 0.6443          & 0.1706          & 0.5950           & 0.3894           & 0.4601           & 0.9181                   \\
                                               & 4.4M                                         & No                                             & 0.6155            & 0.6225          & 0.1693          & 0.5913           & 0.3847           & 0.4543           & 0.9182                   \\ \midrule
\multirow{2}{*}{WavLM B}                    & 4.4M                                         & Yes                                            & 0.7075            & 0.7164          & 0.1732          & \textbf{0.6037}  & \textbf{0.4008}  & \textbf{0.4707}  & \textbf{0.9218}          \\
                                               & 4.4M                                         & No                                             & 0.6717            & 0.6801          & 0.1734          & 0.5981           & 0.3949           & 0.4655           & 0.9186                   \\ \midrule
\multirow{2}{*}{wav2vec 2.0 L}             & 5.4M                                         & Yes                                            & 0.5660            & 0.5808          & 0.1636          & 0.5813           & 0.3718           & 0.4451           & 0.9134                   \\
                                               & 5.4M                                         & No                                             & 0.5559            & 0.5748          & 0.1685          & 0.5731           & 0.3727           & 0.4437           & 0.9155                   \\ \midrule
\multirow{2}{*}{HuBERT L}                  & 5.4M                                         & Yes                                            & 0.6590            & 0.6656          & 0.1718          & 0.5965           & 0.3921           & 0.4635           & 0.9181                   \\
                                               & 5.4M                                         & No                                             & 0.6605            & 0.6637          & 0.1707          & 0.5997           & 0.3950           & 0.4638           & 0.9210                   \\ \midrule
\multirow{2}{*}{WavLM L}                   & 5.4M                                         & Yes                                            & \textbf{0.7114}   & \textbf{0.7183} & \textbf{0.1742} & 0.6034           & 0.4001           & 0.4702           & 0.9213                   \\
                                               & 5.4M                                         & No                                             & 0.6855            & 0.6942          & 0.1714          & 0.5980           & 0.3940           & 0.4647           & 0.9188                   \\ \bottomrule
\end{tabular}
\end{table*}

\subsection{The Results of LibriSQA Part I}
Table~\ref{part1} shows the results of LibriSQA Part I. The term ``ASR pre-training'' refers to whether pre-training is conducted using LibriSpeech in the SQA format. In light of the absence of speech-text multimodal models and the SQA dataset, a direct comparison with previous work is infeasible. To address this, we disorder the model outputs and test them. The results, presented in the last row, reflect the lower bounds of these metrics. By comparing these metrics with others in the preceding rows, the effectiveness of our framework is demonstrated. Remarkably, irrespective of the choice of speech feature extractor or the employment of ASR pre-training, each metric demonstrates significant improvement. Furthermore, it is evident that leveraging ASR pre-training typically yields enhanced model performance. This outcome is reasonable because the ASR pre-training can facilitate the alignment between speech and textual features. However, it is worth noting that even in the absence of such ASR pre-training, requiring the model to autonomously tackle the alignment between speech features and their textual counterparts still produces commendable results (When ASR pre-training is no). Combining this observation with the conclusion in Table~\ref{asr} that SQA is beneficial for ASR tasks, it can be inferred that ASR and SQA are mutually beneficial. Additionally, it is discernible that the choice of feature extractors exerts considerable influence on the outcomes. Examples can be found in Figure~\ref{demo2}. We can see that the model is already capable of understanding speech information, but errors can still occur when filtering for useful information (Bad case).

\begin{figure*}[ht]
\captionsetup[figure]{justification=centering}
\centerline{\includegraphics[width=41pc]{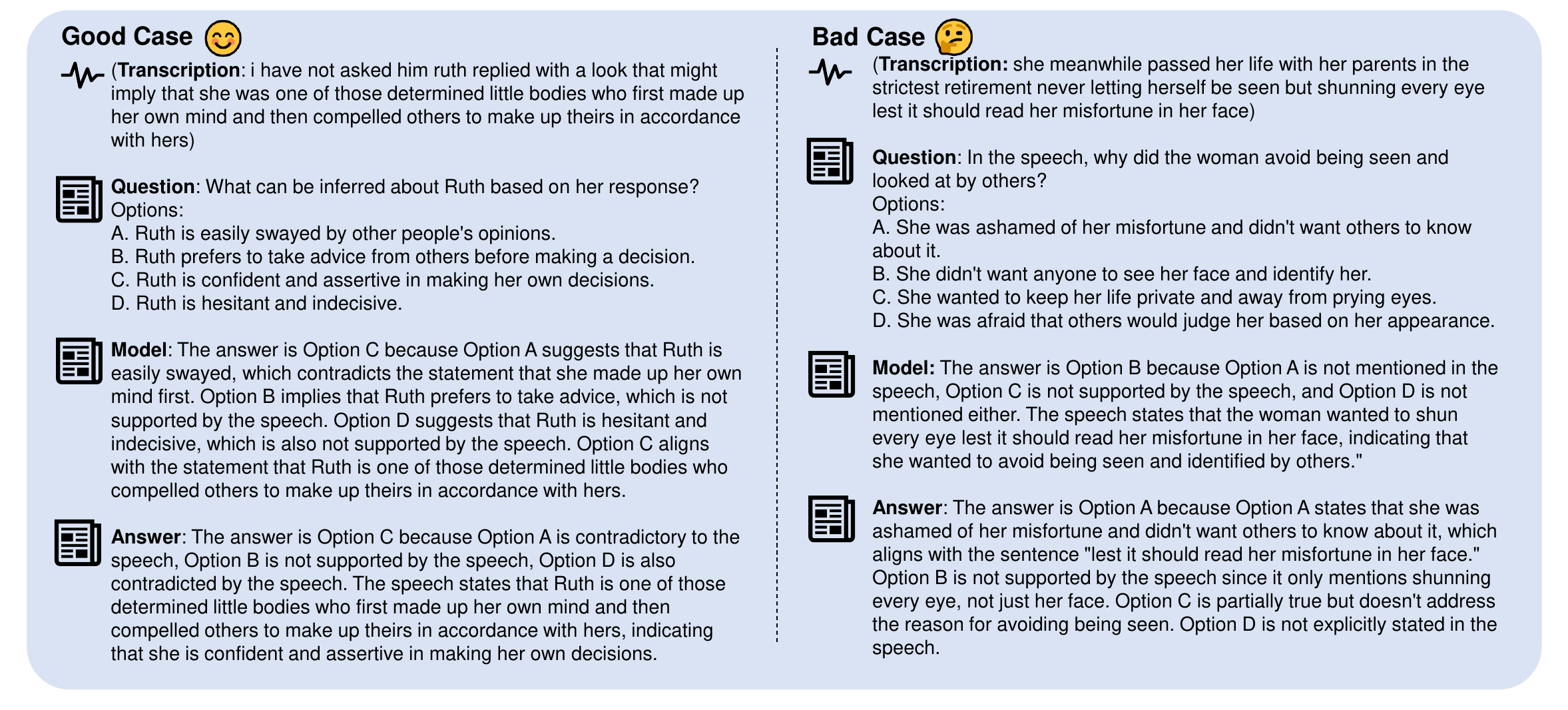}}
\caption{Examples of the model trained with LibriSQA Part II.}
\label{demo3}
\end{figure*}

\subsection{The Results of LibriSQA Part II}

Table~\ref{part2} shows the results of LibriSQA Part II. Here both macro accuracy and F1 are metrics targeted at the options, and we regard them as the most critical benchmarks. The remaining metrics show the analytical effectiveness of the model. The term ``ASR pre-training'' refers to whether pre-training is conducted using LibriSpeech in the SQA format which has been described in Section~\ref{reform}. As shown in the table, we can also see that employing ASR pre-training can boost the performance of the SQA task. Furthermore, it is evident that the choice of speech feature extractors impacts the outcomes significantly, which is consistent with the results of Table~\ref{part1}. Among them, WavLM emerges as the most effective, and the ``large'' models consistently outperform their ``base'' counterparts. We suppose that this might be due to WavLM's capacity for universal representations, whereas wav2vec 2.0 and HuBERT appear more tailored towards the ASR task. Examples can be found in Figure~\ref{demo3}. We can see the ability of the model to make reasonable inferences directly from the speech document, rather than only answering the words contained in the speech document. This capability is made possible by the improvements we have made to both the dataset and the model compared to previous work.

\begin{figure}
    \begin{tikzpicture}
\begin{axis}[
    xlabel={SNR (dB)},
    ylabel={Accuracy (\%)},
    width=8cm,
    height=6cm,
    xmin=-50, xmax=20,
    ymin=40, ymax=80,
    legend pos=north west,
    ymajorgrids=true,
    grid style=dashed,
    label style={font=\small},           % X轴和Y轴的标签字体大小
    tick label style={font=\small},       % 刻度标签字体大小
    legend style={font=\small},
    ylabel near ticks,
    grid style={dashed,gray!30},
    major grid style={lightgray},
    legend pos=north west,
]

% Line 1
\addplot[
    color=blue,
    mark=square*,
    mark size=1.5pt,
    smooth,
    ] 
    coordinates {
    (-55,43.23)(-50,43.25)(-45,43.15)(-40,43.40)(-35,43.80)(-30,44.80)(-25,46.53)(-20,51.53)(-15,60.99)(-10,68.24)(-5,70.50)(0,71.53)(5,71.60)(10,71.68)(15,71.70)(20,71.65)
    };

% Line 2
\addplot[
    color=red,
    mark=triangle*,
    mark size=1.5pt,
    smooth,
    ] 
    coordinates {
    (-50,43.22)(-50,43.22)(-45,43.18)(-40,43.19)(-35,43.21)(-30,43.20)(-25,43.67)(-20,47.67)(-15,56.26)(-10,65.34)(-5,70.10)(0,74.27)(5,74.90)(10,75.04)(15,75.06)(20,75.03)
    };

\legend{End-to-end, Cascade}

\end{axis}
\end{tikzpicture}
\caption{Comparison between cascade and end-to-end methods under different noise levels. SNR is short for signal-to-noise ratio.}
\label{compare}
\end{figure}
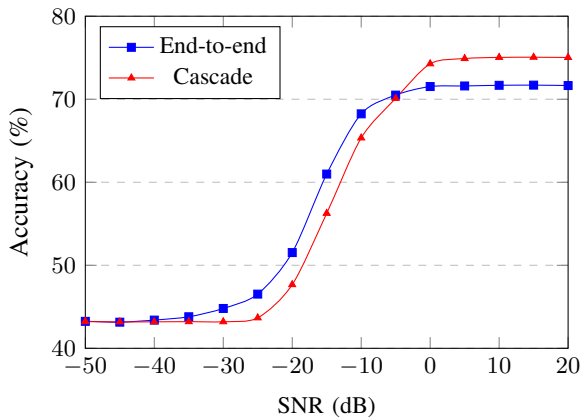

\subsection{Comparison with Cascade Models}

\begin{table*}[]
\centering
\caption{The impact of fine-tuned speech feature extractors on LibriSpeech. The results are conducted on LibriSQA Part II. Pre-trained means the pre-trained version that didn't use labeled text data. Fine-tuned means the fine-tuned version that uses labeled text data}
\label{fttable}
\begin{tabular}{c|cccccccc}
\toprule
Speech Feature Extractor                            & Version  & Accruacy$\uparrow$ & F1$\uparrow$     & BLEU-1$\uparrow$ & ROUGE-1$\uparrow$ & ROUGE-2$\uparrow$ & ROUGE-L $\uparrow$& BERT Similarity$\uparrow$ \\ \midrule
\multirow{2}{*}{wav2vec 2.0 Base}  & Pre-trained & \textbf{0.5846}   & \textbf{0.5943} & \textbf{0.1695} & \textbf{0.5934}  & \textbf{0.3927}  & \textbf{0.4621}  & \textbf{0.9167}          \\
                                   & Fine-tuned       & 0.5771   & 0.5840 & 0.1690 & 0.5897  & 0.3823  & 0.4516  & \textbf{0.9167}          \\ \midrule
\multirow{2}{*}{wav2vec 2.0 Large} & Pre-trained & 0.5660   & 0.5808 & 0.1636 & 0.5813  & 0.3718  & 0.4451  & 0.9134          \\
                                   & Fine-tuned       & \textbf{0.5684}   & \textbf{0.5821} & \textbf{0.1677} & \textbf{0.5857}  & \textbf{0.3775}  & \textbf{0.4492}  & \textbf{0.9159}          \\ \midrule
\multirow{2}{*}{HuBERT Large}      & Pre-trained & \textbf{0.6590}   & 0.6656 & 0.1718 & 0.5965  & 0.3921  & 0.4635  & 0.9181          \\
                                   & Fine-tuned       & 0.6547   & \textbf{0.6700} & \textbf{0.1735} & \textbf{0.6024}  & \textbf{0.3997}  & \textbf{0.4696}  & \textbf{0.9188}          \\ \bottomrule
\end{tabular}
\end{table*}

A common question about SQA is why not employ an ASR model to transcribe the speech into text directly and then proceed with text-based QA. Although this approach is feasible, we still contend that training an end-to-end SQA model is meaningful. We support our stance with the following reasons:

%1) Figure \ref{compare} presents a comparison between the results of the end-to-end model and the cascade method under various noise conditions. In the cascade method, the ASR module we use is the whisper-tiny model \cite{whisper}. It shows that under low-noise conditions (i.e. high SNR), the cascade method yields better results. However, in high-noise environments, the end-to-end approach is superior, suggesting our method may offer enhanced robustness. This observation aligns with the findings in \cite{speechbert}. Although advancements in ASR could potentially boost the performance of the cascade method, Table \ref{part1} and \ref{part2} also demonstrate that speech feature extractors play a pivotal role in influencing the results.
1) Figure~\ref{compare} {presents a comparison between the results of the end-to-end model and the cascade method under various noise conditions. In the cascade method, we use the whisper-tiny model }\cite{whisper} {as our ASR module. The Cascade method has 44M trainable parameters, while our end-to-end model only has 5.4M trainable parameters. The comparison shows that the cascade method performs better under low-noise conditions (i.e., high SNR). However, in high-noise environments, the end-to-end approach is superior, suggesting that our method may offer better robustness. This observation aligns with the findings in }\cite{speechbert}. {When the noise is very high, the performance of both models converges at around 43\%. We believe this corresponds to the situation where no model can extract any effective information from the speech or ASR text, leading to judgments being made directly from the questions and options. Although advancements in ASR could potentially improve the performance of the cascade method, }Table~\ref{part1} and \ref{part2} {also demonstrate the significant influence of speech feature extractors on the results. Therefore it is also possible to improve the performance of the end-to-end method by replacing the speech feature extractor.}

%2) The cascade method risks losing the prosodic information inherent in speech, which could be vital for certain questions, such as determining a speaker's emotion \cite{m1,m2}. The loss of such information might hinder the model's ability to answer related questions accurately.

2) The cascade method runs the risk of losing the prosodic information present in speech, which can be crucial for certain inquiries, such as identifying a speaker's emotion \cite{m1,m2}. The absence of this information could potentially impede the model's accuracy in answering related questions.

%3) Relying on the cascade method results in the total loss of audio information, effectively stripping the model of its capacity to discern some questions like inquiries about the speaker's surrounding environment.
3) When relying on the cascade method, the model loses all audio information, which means it cannot effectively answer certain questions, such as inquiries about the speaker's surrounding environment.

%4) An analysis of Table \ref{asr}-\ref{part2} further corroborates that the end-to-end approach can effectively integrate both ASR and SQA tasks. By adopting a unified text generation training methodology, this lays the groundwork for the development of comprehensive, general-purpose multimodal models.

4) An analysis of Table~\ref{asr}-\ref{part2} further confirms that the end-to-end approach can successfully combine both ASR and SQA tasks. Using a unified text generation training methodology establishes the foundation for building comprehensive, versatile multimodal models.

% \usepackage{multirow}

\subsection{Impact of Fine-Tuned Feature Extractors on LibriSpeech}
\label{ftextractor}

{As mentioned in} Section~\ref{extractor}, {it is possible that the speech feature extractor was fine-tuned for ASR tasks on LibriSpeech which is the speech source we use. This raises the potential issue of text annotation leakage. However, we circumvented this problem in all our experiments above by choosing the pre-trained version of the speech feature extractor rather than the fine-tuned version. Nonetheless, whether choosing the fine-tuned version of the feature extractor truly impacts the results remains to be experimentally verified. In this section, we select wav2vec 2.0 Base, wav2vec 2.0 Large, and HuBERT Large for our experiments. The reason for this selection is that only these models have official fine-tuned versions for LibriSpeech available on Hugging Face }\cite{huggingface}, {whereas other versions of the feature extractors do not. The results, displayed in} Table~\ref{fttable}, {indicate that the fine-tuned versions of the speech feature extractors do not exhibit significant advantages over the pre-trained versions. In fact, there is a slight decline in performance with wav2vec 2.0 Base. These results suggest that there was no leakage of text labels. We believe this is because we only implicitly use label information and do not directly employ the text output from the fine-tuned model. Specifically, we use the output of the transformer layers from the fine-tuned model, not the linear layers, as features. Since the earlier layers in fine-tuning learn more generic features and the later layers learn more task-specific features, it is likely that generic features are less prone to leaking text information, thereby not influencing the model's performance. Moreover, the results also validate the viability of using pre-trained versions of speech feature extractors, consistent with the effective outcomes we observed in} Tables \ref{asr}, \ref{part1}, and \ref{part2}, {where we used the pre-trained versions.}

\subsection{Computation Cost Analysis}
{
The total number of parameters in our model is approximately 7 billion, with at least 4.4 million parameters being trainable and requiring around 16.78 MB of storage space. Utilizing the original configuration of the LLaMA-adapter }\cite{llamaadapter},{ where parameters are still in 32-bit without employing other methods for reducing memory usage or accelerating the process, we trained the model on four A100 GPUs. Training for one epoch on the proposed LibriSQA dataset required 4 hours. For inference, we used a single A100 GPU, and testing on the LibriSQA dataset took 4 hours.}
\section{Conclusion and future work}

%In this paper we present LibriSQA, a novel free-form and open-ended SQA dataset for LLMs. We have also present a new framework for integrating speech and text. Finally we examine our framework using LibriSpeech and LibriSQA and get promising results.

In this paper, we introduce LibriSQA, a novel dataset for free-form and open-ended SQA which is designed to facilitate the research in the perception and understanding of speech leveraging large language models. In addition, we propose a novel framework for integrating speech and text. We evaluate our framework using both SQA and ASR tasks and obtain promising results. In the future, we will consider expanding the scope of the dataset by adding more audio-related questions to enhance its generality.

%\section*{Appendix}

%Appendixes, if needed, appear before the acknowledgment.

\section*{References}
\def\refname{}

\end{document}